\documentclass[runningheads]{llncs}
\usepackage{graphicx}
\usepackage{amsmath,amssymb}
\usepackage{epsfig}
\usepackage{booktabs}
\usepackage{enumerate}
\usepackage{subfigure}
\usepackage{caption}
\usepackage{mathtools}
\usepackage{enumitem}
\usepackage{placeins}
\usepackage{graphicx}
\usepackage{url}

\usepackage{color}
\usepackage[width=122mm,left=12mm,paperwidth=146mm,height=193mm,top=12mm,paperheight=217mm]{geometry}
\usepackage{float}
\usepackage{subfig}

\DeclareMathOperator{\E}{\mathbb{E}}
\begin{document}
\def\mA{\mathcal{A}}
\def\mB{\mathcal{B}}
\def\mC{\mathcal{C}}
\def\mD{\mathcal{D}}
\def\mE{\mathcal{E}}
\def\mF{\mathcal{F}}
\def\mG{\mathcal{G}}
\def\mH{\mathcal{H}}
\def\mI{\mathcal{I}}
\def\mJ{\mathcal{J}}
\def\mK{\mathcal{K}}
\def\mL{\mathcal{L}}
\def\mM{\mathcal{M}}
\def\mN{\mathcal{N}}
\def\mO{\mathcal{O}}
\def\mP{\mathcal{P}}
\def\mQ{\mathcal{Q}}
\def\mR{\mathcal{R}}
\def\mS{\mathcal{S}}
\def\mT{\mathcal{T}}
\def\mU{\mathcal{U}}
\def\mV{\mathcal{V}}
\def\mW{\mathcal{W}}
\def\mX{\mathcal{X}}
\def\mY{\mathcal{Y}}
\def\mZ{\mathcal{Z}}

\def\1n{\mathbf{1}_n}
\def\0{\mathbf{0}}
\def\1{\mathbf{1}}

\def\A{{\bf A}}
\def\B{{\bf B}}
\def\C{{\bf C}}
\def\D{{\bf D}}
\def\E{{\bf E}}
\def\F{{\bf F}}
\def\G{{\bf G}}
\def\H{{\bf H}}
\def\I{{\bf I}}
\def\J{{\bf J}}
\def\K{{\bf K}}
\def\L{{\bf L}}
\def\M{{\bf M}}
\def\N{{\bf N}}
\def\O{{\bf O}}
\def\P{{\bf P}}
\def\Q{{\bf Q}}
\def\R{{\bf R}}
\def\S{{\bf S}}
\def\T{{\bf T}}
\def\U{{\bf U}}
\def\V{{\bf V}}
\def\W{{\bf W}}
\def\X{{\bf X}}
\def\Y{{\bf Y}}
\def\Z{{\bf Z}}

\def\a{{\bf a}}
\def\b{{\bf b}}
\def\c{{\bf c}}
\def\d{{\bf d}}
\def\e{{\bf e}}
\def\f{{\bf f}}
\def\g{{\bf g}}
\def\h{{\bf h}}
\def\i{{\bf i}}
\def\j{{\bf j}}
\def\k{{\bf k}}
\def\l{{\bf l}}
\def\m{{\bf m}}
\def\n{{\bf n}}
\def\o{{\bf o}}
\def\p{{\bf p}}
\def\q{{\bf q}}
\def\r{{\bf r}}
\def\s{{\bf s}}
\def\t{{\bf t}}
\def\u{{\bf u}}
\def\v{{\bf v}}
\def\w{{\bf w}}
\def\x{{\bf x}}
\def\y{{\bf y}}
\def\z{{\bf z}}

\def\balpha{\mbox{\boldmath{$\alpha$}}}
\def\bbeta{\mbox{\boldmath{$\beta$}}}
\def\bdelta{\mbox{\boldmath{$\delta$}}}
\def\bgamma{\mbox{\boldmath{$\gamma$}}}
\def\blambda{\mbox{\boldmath{$\lambda$}}}
\def\bsigma{\mbox{\boldmath{$\sigma$}}}
\def\btheta{\mbox{\boldmath{$\theta$}}}
\def\bomega{\mbox{\boldmath{$\omega$}}}
\def\bxi{\mbox{\boldmath{$\xi$}}}
\def\bnu{\mbox{\boldmath{$\nu$}}}                                  
\def\bphi{\mbox{\boldmath{$\phi$}}}

\def\bDelta{\mbox{\boldmath{$\Delta$}}}
\def\bOmega{\mbox{\boldmath{$\Omega$}}}
\def\bPhi{\mbox{\boldmath{$\Phi$}}}
\def\bLambda{\mbox{\boldmath{$\Lambda$}}}
\def\bSigma{\mbox{\boldmath{$\Sigma$}}}
\def\bGamma{\mbox{\boldmath{$\Gamma$}}}

\newcommand{\myminimum}[1]{\mathop{\textrm{minimum}}_{#1}}
\newcommand{\mymaximum}[1]{\mathop{\textrm{maximum}}_{#1}}    
\newcommand{\mymean}[1]{\mathop{\textrm{mean}}_{#1}}
\newcommand{\myvar}[1]{\mathop{\textrm{Variance}}_{#1}}
\newcommand{\mymin}[1]{\mathop{\textrm{minimize}}_{#1}}
\newcommand{\mymax}[1]{\mathop{\textrm{maximize}}_{#1}}
\newcommand{\mymins}[1]{\mathop{\textrm{min.}}_{#1}}
\newcommand{\mymaxs}[1]{\mathop{\textrm{max.}}_{#1}}  
\newcommand{\myargmin}[1]{\mathop{\textrm{argmin}}_{#1}} 
\newcommand{\myargmax}[1]{\mathop{\textrm{argmax}}_{#1}} 
\newcommand{\myst}{\textrm{s.t. }}

\newcommand{\denselist}{\itemsep -1pt}
\newcommand{\sparselist}{\itemsep 1pt}

\definecolor{pink}{rgb}{0.9,0.5,0.5}
\definecolor{purple}{rgb}{0.5, 0.4, 0.8}   
\definecolor{gray}{rgb}{0.3, 0.3, 0.3}
\definecolor{mygreen}{rgb}{0.2, 0.6, 0.2}

\newcommand{\cyan}[1]{\textcolor{cyan}{#1}}
\newcommand{\red}[1]{\textcolor{red}{#1}}  
\newcommand{\blue}[1]{\textcolor{blue}{#1}}
\newcommand{\magenta}[1]{\textcolor{magenta}{#1}}
\newcommand{\pink}[1]{\textcolor{pink}{#1}}
\newcommand{\green}[1]{\textcolor{green}{#1}} 
\newcommand{\gray}[1]{\textcolor{gray}{#1}}    
\newcommand{\mygreen}[1]{\textcolor{mygreen}{#1}}    
\newcommand{\purple}[1]{\textcolor{purple}{#1}}       

\definecolor{greena}{rgb}{0.4, 0.5, 0.1}
\newcommand{\greena}[1]{\textcolor{greena}{#1}}

\definecolor{bluea}{rgb}{0, 0.4, 0.6}
\newcommand{\bluea}[1]{\textcolor{bluea}{#1}}
\definecolor{reda}{rgb}{0.6, 0.2, 0.1}
\newcommand{\reda}[1]{\textcolor{reda}{#1}}

\newcommand{\mtodo}[1]{{\color{red}$\blacksquare$\textbf{[TODO: #1]}}}
\newcommand{\myheading}[1]{\vspace{1ex}\noindent \textbf{#1}}

\def\changemargin#1#2{\list{}{\rightmargin#2\leftmargin#1}\item[]}
\let\endchangemargin=\endlist
                                               
\newcommand{\cm}[1]{}

\def\xbi{\overline{\x}_i}
\def\wbi{\overline{\w}_{(i)}}
\def\wb{\overline{\w}}
\def\Ib{\overline{\I}}
\def\invC{\C^{-1}}
\def\invCi{\C_{(i)}^{-1}}
\def\ab{\overline{\balpha}}
\def\abi{\overline{\balpha}_{(i)}}
\def\Kb{\overline{\K}}
\def\Xb{\overline{\X}}
\def\kbi{\overline{\k}_{i}}
\def\Kzz{\K_{\z\z}}
\def\Kzx{\K_{\z\x}}
\def\Xsub{\X_{sub}}
\def\ssub{\s_{sub}}
\def\wbsub{\overline{\w}_{sub}}
\def\dsub{\d_{sub}}
\def\invCsub{\C^{-1}_{sub}}
\def\etal{\emph{et al}.}
\def\DS{\textcolor{red}}
\newcommand{\norm}[1]{\left\lVert#1\right\rVert}

\pagestyle{headings}
\mainmatter
\def\ECCV18SubNumber{2788}  
\def\subFigSzab{\linewidth}
\title{A+D Net: Training a Shadow Detector with \\ Adversarial Shadow Attenuation} 

\titlerunning{Shadow Detection with Adversarial Shadow Attenuation}

\authorrunning{Le et al.}

\author{Hieu Le \inst{1}\and  Tomas F. Yago Vicente  \inst{1,2} \and Vu Nguyen \inst{1} \and \\ Minh Hoai \inst{1} \and Dimitris Samaras \inst{1} \\
\texttt{\small \{hle, tyagovicente, vhnguyen, \\ minhhoai, samaras\} @cs.stonybrook.edu}}
\institute{Stony Brook University, Stony Brook, NY 11794, USA \and Amazon/A9}



\maketitle

\begin{abstract}
 We propose a novel GAN-based framework for detecting shadows in images, in which a shadow detection network (D-Net) is trained together with a shadow attenuation network (A-Net) that generates adversarial training examples. The A-Net modifies the original training images constrained by a simplified physical shadow model and is focused on fooling the D-Net's shadow predictions. Hence, it is effectively augmenting the training data for D-Net with hard-to-predict cases. The D-Net is trained to predict shadows in both original images and generated images from the A-Net. Our experimental results show that the additional training data from A-Net significantly improves the shadow detection accuracy of D-Net. Our method outperforms the state-of-the-art methods on the most challenging shadow detection benchmark (SBU) and also obtains state-of-the-art results on a cross-dataset task, testing on UCF. Furthermore, the proposed method achieves accurate real-time shadow detection at 45 frames per second.
\end{abstract}

\section{Introduction}

  
  
   

Shadows occur frequently in natural scenes, and can hamper many tasks such image segmentation, object tracking, and semantic labeling.
Shadows are formed in complex physical interactions between light sources, geometry and materials of the objects in the scene. Information about the physical environment such as sparse 3D scene reconstructions \cite{snavely_2015_shadows}, rough geometry estimates \cite{Panagopoulos13}, and multiple images of the same scene under different illumination conditions \cite{Sunkavalli:2007:FTV} can aid shadow detection. Unfortunately,  inferring the physical structure of a general scene from a single image is still a difficult problem.  



\begin{figure}[h]
	\centering
  \includegraphics[width=\textwidth]{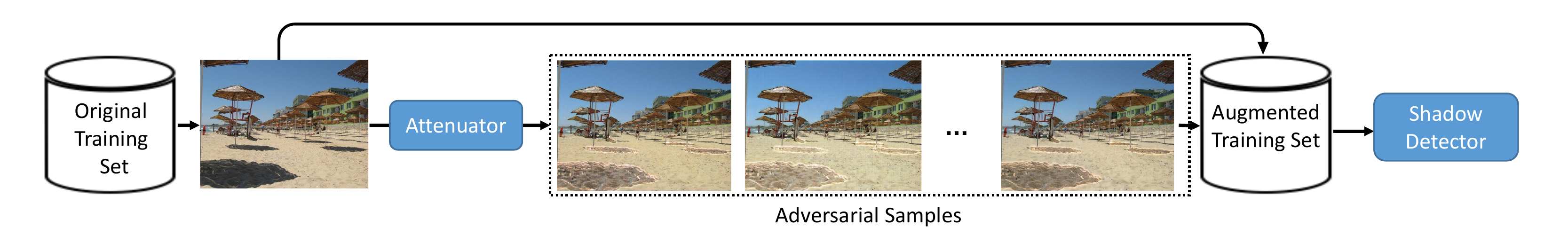}
    \vskip -0.1in
  \caption{\textbf{Adversarial shadow attenuation.} The attenuator takes an original shadow image and generates different adversarial shadow samples to train the shadow detector.}
  \label{fig:Teaser}
\end{figure}


The difficulty of shadow detection is exacerbated when dealing with consumer-grade photographs and web images~\cite{Lalonde10}. Such images often come from non-linear camera sensors, and present many compression and noise artifacts. In this case, it is better to train and use appearance-based classifiers~\cite{Zhu10,guo11,Vicente-et-al-ICCV15,Khan_2014_CVPR} rather than relying on physical models of illumination~\cite{Finlayson06,Finlayson09}.  Shadow classifiers, however, require annotated training data, and the performance of a classifier often correlates with the amount of training data. Unfortunately, annotated shadow data is expensive to collect and label. Only recently  available training data has increased from a few hundred images~\cite{guo11,Zhu10} to a few thousands~\cite{Vicente-etal-ECCV16} thus enabling  training  more powerful shadow classifiers based on deep convolutional neural networks~\cite{Vicente-etal-ECCV16,VuICCV2017}. Nevertheless, even a few thousand images is a tiny amount compared to datasets that have driven progress in deep learning \cite{imagenet_cvpr09,coco}. It is therefore safe to assume that the performance of deep learning shadow classifiers has not saturated yet, and it can be improved with more training data. Unfortunately, collecting and annotating shadow data is a laborious process. Even a lazy annotation approach \cite{Vicente-et-al-CVPR16} takes significant effort; the annotation step itself takes 20 seconds per image, not including data collection and cleansing efforts. 

In this paper, instead of collecting additional data, we propose a method to increase the utility of available shadow data to the fullest extent. The main idea is to generate a set of augmented training images from a single shadow image by weakening the shadow area in the original training image. We refer to this process as shadow attenuation and we train a deep neural network to do so, called A-Net. This network modifies original shadow images so as to weaken the shadow effect, as illustrated in Fig. \ref{fig:Teaser}. The generated images serve as additional challenging training samples for a shadow detector D-Net. We present a novel framework, where the shadow attenuator and the shadow detector are trained jointly in an adversarial manner. The output of the attenuation model A-Net provides adversarial training samples with harder-to-detect shadow areas to improve the overall reliability of the detector D-Net. 


Recent research also suggests that deep networks are highly sensitive to adversarial perturbations~\cite{MoosaviDezfooli2016DeepFoolAS,tramer2018ensemble,xie2017adversarial}. By jointly training A-Net and D-Net, we directly enhance the resistance of the detector D-Net to adversarial conditions and improve the generalization of the detector,  following the recent trend ~\cite{zhang2018mixup,Erraqabi2018A3TAA,Volpi_2018_CVPR}.

Essentially, what is being proposed here is a data augmentation method for shadow detection. It is different from other data augmentation methods, and it does not suffer from two inherent problems of general data augmentation approaches, which are: 1) the augmented data might be very different from the real data, having no impact on the generalization ability of the trained classifier on real data, and 2) it is difficult to ensure that the augmented data samples have the same labels as the original data, and this leads to training label noise. A popular approach to address these problems is to constrain the augmented data samples to be close to the original data, e.g., setting an upper bound for the $L_2$ distance between the original sample and the generated sample. However, it is difficult to set the right bound; a big value would create label noise while a small value would produce augmented samples that are too similar to the original data, yielding no benefit. In this paper, we address these two problems in a principled way, specific to shadow detection. Our idea is to use a physics model of shadows and illumination to guide the data generation process and to estimate the probability of having label noise.



Note that we aim to attenuate the shadow areas, not to remove them. Shadow removal is an important problem, but training a good shadow removal network would require many training pairs of corresponding shadow/shadow-free images, which are not available. Furthermore, completely removed shadows would correspond to having label noise, and this might hurt the performance of the detector.

Experimental results show that our shadow detector outperforms the state-of-the-art methods in the challenging shadow detection benchmark SBU~\cite{Vicente-etal-ECCV16} as well as on the cross-dataset task (training on SBU and testing on the UCF dataset~\cite{Zhu10}). Furthermore, our method is more  efficient than many existing ones because it does not require a post-processing step such as patch averaging or conditional random field (CRF) smoothing. Our method detects shadows at 45 frames per second for $256\times256$ input images.

\section{Related Work}
Single image shadow detection is a well studied problem. Earlier work focused on physical modeling of illumination~\cite{Finlayson09,Finlayson06}. These methods render illumination invariant representations of the images where shadow detection is trivial. These methods, however, only work well for high quality images taken with narrow-band sensors~\cite{Lalonde10}. Another early attempt to incorporate physics based constraints with rough geometry was the approach of Panagopoulos~\etal ~\cite{5206665} where the illumination environment is modeled as a mixture of von Mises-Fisher distributions~\cite{Banerjee} and the shadow pixels are segmented via a graphical model.
Recently, data-driven approaches based on learning classifiers~\cite{guoPami,huang11,Vicente-et-al-ICCV15,Khan_2014_CVPR} from small annotated datasets~\cite{Zhu10,guo11} have shown more success. For instance, Vicente \etal~\cite{Vicente-et-al-ICCV15,Vicente-etal-PAMI18} optimized a multi-kernel Least-Squares SVM based on leave-one-out estimates. This approach yielded accurate results on the UCF~\cite{Zhu10} and UIUC~\cite{guo11} datasets, but its underlying training procedure and optimization method cannot handle a large amount of training data.

To handle and benefit from a large amount of training data, recent shadow detection methods have been developed based on the stochastic gradient descent training of deep neural networks. Vicente \etal~\cite{Vicente-etal-ECCV16} proposed a stacked-CNN architecture, combining an image-level Fully Convolution Neural Network (FCN) with a patch-CNN. This approach achieved good detection results, but it is cumbersome as the Fully Connected Network (FCN) has to be trained before its predictions are used to train the patch-CNN. Similarly, testing was computationally expensive as it requires the FCN prediction followed by predictions of densely sampled patches covering the testing image. Recently, Nguyen~\etal~\cite{VuICCV2017} presented scGAN, a method based on Generative Adversarial Networks (GANs)~\cite{GoodfellowPMXWOCB14}. They proposed a parametric conditional GAN~\cite{mirza2014conditional} framework, where the generator was trained to generate the shadow mask, conditioned on an input RGB patch and a sensitivity parameter. To obtain the final shadow mask for an input image, the generator must be run on multiple image patches at multiple scales and the outputs are averaged. Their method achieved good results on the SBU dataset, but the detection procedure was computationally expensive at test time. 
Our proposed method also uses adversarial training for shadow detection, but it is fundamentally different from scGAN. scGAN uses the generator to generate a binary {\em shadow mask} conditioned on the input image, while our method uses the generator to generate augmented training images in RGB space.  Furthermore, while scGAN uses the discriminator as a regulator to encourage global consistency, the discriminator in our approach plays a more prominent role for shadow pixel classification. 
In contrast to scGAN, our method does not require post processing or output averaging, leading to real-time shadow detection. Another method that uses GAN for shadow detection is Stacked Conditional GAN~\cite{Wang_2018_CVPR}. This method, however, requires the availability of shadow-free images. Another recent approach~\cite{Hu_2018_CVPR} proposes to use contextual information for a better shadow detection. Contextual information is incorporated by  having several spatial-directional recurrent neural networks. While this method yields excellent results on shadow detection benchmarks, it also requires running a CRF as a post-processing step.

 \begin{figure*}[t]
 \centering
  \includegraphics[width = 0.8\textwidth]{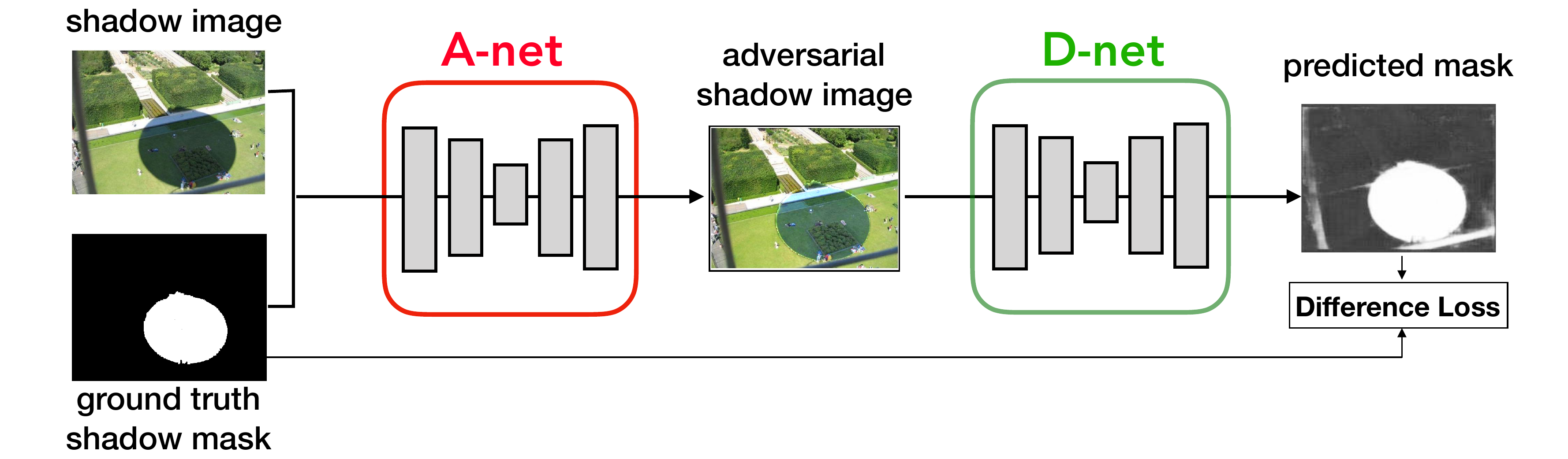}
  \vskip -0.1in
  \caption{{\bf Adversarial training of a shadow detector.} A-Net takes a shadow image and its corresponding shadow mask as input, and generates an adversarial example by attenuating the shadow regions in the input image. The attenuated shadows are less discernible and therefore harder to detect. D-Net takes this image as input and aims to recover the original shadow mask.}
  \label{framework}
\end{figure*}

We propose a method to improve shadow detection with augmented training examples, in sync  with recent  trends on data augmentation. For example, Zhang \etal~\cite{zhang2018mixup} proposed a simple augmentation method by enriching the dataset with the linear combinations of pairs of examples and their labels to improve the generalization of the network and its resistance toward adversarial examples. Another approach that used  adversarial examples for training a network was proposed by Shrivastava~\etal~\cite{appleShrivastavaPTSW16}. They adversarially trained a Refiner network that inputs synthetic examples and outputs more realistic images. The refined examples can be used as additional training data. In a similar way, our proposed Attenuator (A-Net) takes original training images and generates realistic images with attenuated shadows that act as additional training examples for our shadow detector. The generation of adversarial examples is an integral part of the joint training process with the detector (D-Net), in contrast to \cite{appleShrivastavaPTSW16} where the generated data is a preprocessing step to enrich the training set. The effects of the shadow Attenuator can also be seen as related to adversarial perturbations \cite{Moosavi-Dezfooli_2017_CVPR}: A-Net modifies the input images so as to fool the predictions of the shadow detector D-Net. Adversarial examples also can be used to improve the generalization of the network for domain adaptation \cite{Volpi_2018_CVPR} in which a conditional GAN is used to perform feature augmentation.

\section{Adversarial Training and Attenuation}

\subsection{Framework Overview}
We present a novel framework for shadow detection based on adversarial training and shadow attenuation. Our proposed model contains two jointly trained deep networks. Fig. \ref{framework} illustrates the flow diagram of our framework. The shadow attenuation network, called Attenuator or A-Net, takes as input a shadow image and its corresponding shadow mask. Based on these inputs, the Attenuator generates a version of the input image where the shadows have been attenuated. Attenuation can be thought of as partial shadow removal. The image generated by the Attenuator is  fed into a shadow detection network, called Detector or D-Net, which predicts the shadow areas. On each training iteration, D-Net also takes the original input image, and learns to predict the corresponding annotated ground-truth shadow mask. 

A-Net is trained to attenuate shadow regions so as to fool the shadow detector. In particular, for pixels inside the provided shadow mask, A-Net manipulates the values of the pixels to disguise them as non-shadow pixels so that they cannot be recognized by D-Net. 
 We further constrain the attenuation transformation using  a loss that incorporates physics-inspired shadow domain knowledge. This enhances the quality of the generated pixels, improving the generalizability of the detector. At the same time, A-Net learns not to change the values or the pixels outside the shadow mask. We enforce this with a loss that penalizes the difference between the generated image and the input image on the area outside of the shadow mask (non-shadow pixels). The adversarial training process with all the aforementioned constraints and the back propagation error from the shadow detection network guides A-Net to perform shadow attenuation.

The detector network, D-Net, takes the adversarial examples generated by A-Net and predicts shadow masks. Shadow areas in the images generated by A-Net are generally harder to detect than in the input images, since A-Net is trained to attenuate the shadows to fool D-Net. As a result, D-Net is trained with challenging examples in addition to the original training examples. As D-Net improves its ability to detect shadows, A-Net  also improves its ability to attenuate shadows to confound D-Net with tougher adversarial examples. This process strengthens the shadow detection ability of D-Net.




\subsection{Physics-based Shadow and Illumination Model} \label{sec:illumModel}

We use a physics-based illumination model to guide the data generation process and  avoid  label noise. We use the simplified illumination model used by Guo \etal~\cite{guo11,guoPami} where, each pixel is lit by a combination of direct and environment lights: 
$\label{eq:pareto mle2}
{I_i} = \left(k_i L_d  + L_e\right)R_i,
$
where $I$ is an image and $I_i$ denotes the color of the $i^{th}$ pixel of the image. $R_i$ is the surface reflectance corresponding to the $i^{th}$ pixel. $L_d$ and $L_e$ are $3\times 1$ vectors representing the colors and intensities of  the direct light and the environment light (which models area sources and inter reflections), respectively. $k_i\in [0,1]$ is the shadowing factor that indicates how much of the direct light reaches the pixel $i$. $k_i$ remains close to 0 for the umbra region of the shadow, while it gets increasingly close to 1 in the penumbra region. For pixels inside shadow-free areas $k_i = 1$. We can relate the original shadow region and its corresponding shadow-free version by the ratio:
\begin{equation*}
\frac{{I_i}^{\textrm{shadow-free}}}{{I_i}^{\textrm{shadow}}} = \frac{L_d + L_e}{k_i  L_d  + L_e}.
\end{equation*}

By taking the ratio between the shadow-free and in-shadow values, we have eliminated the unknown reflectance factor. We assume that the direct light is constant over the scene depicted by the image, and the effects of the environment light are similar for all pixels. We incorporate this model into the training process of both A-Net and D-Net:
\begin{itemize}
\item \textbf{A-Net:} We design the \textsl{physics loss} to enforce the illumination ratios for pixels inside an attenuated shadow area to have a small variance.
\item \textbf{D-Net:} We directly estimate the illumination ratio between the areas inside and outside the shadow mask to measure shadow strength in the attenuated images to avoid training label noise. 
\end{itemize}



\subsection{A-Net: Shadow Attenuator Network}
\label{Sec:A-Net}
\begin{figure}[t]
\centering
  \includegraphics[width=0.75\textwidth]{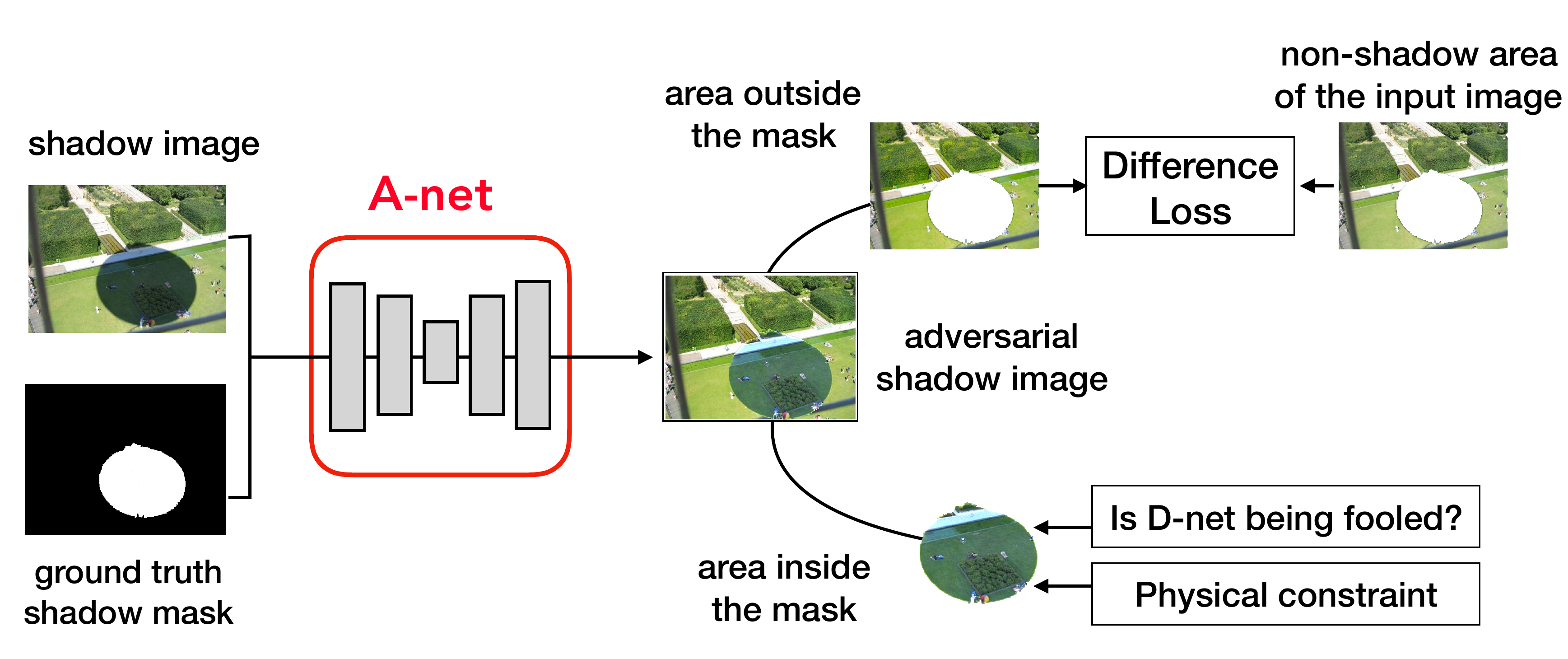}
  \vskip -0.1in 
  \caption{{\bf A-Net.}  The area outside the shadow mask is constrained by the difference loss with respect to the input image. The area inside the shadow mask is constrained by the feedback from D-Net and the physics based constraint.}
  \label{fig:Anet}
\end{figure}

The shadow attenuator network A-Net is trained to re-illuminate only the shadow areas  so that they cannot be detected by the detector network D-Net. To obtain useful and realistic attenuated shadows, A-Net aims to fool D-Net while respecting a physical illumination model. Fig. \ref{fig:Anet} shows the training process of A-Net, which attenuates shadow areas under the following constraints and objectives: 
1) Values of non-shadow pixels  are preserved.
2) Shadow pixels  are re-illuminated such that D-Net cannot recognize them as shadow pixels.
3) The resulting pixel transformation obeys physics-inspired illumination constraints.

These constraints and objectives can be incorporated in the training of A-Net by defining a proper loss function.  Let $I$ denote an input image, and $M(I)$ be the shadow mask of $I$. Let $A(I)$ denote the output of A-Net for the input pair of $I$ and $M(I)$ (here we write $A(I)$ as the short form for $A(I, M(I))$). Let $D(I)$ denote the output of D-Net for an input image $I$, i.e.  the predicted shadow mask. Ideally, the output should be~1 for shadow pixels and 0 otherwise. The objective of A-Net's training is to minimize a weighted combination of three losses:
\begin{align}
\mathcal{L}_A(I) =
  \lambda_{nsd}\mathcal{L}_{nsd}(I) + \lambda_{sd}\mathcal{L}_{sd}(I) + \lambda_{ph}\mathcal{L}_{ph}(I), 
\end{align}
where $\mathcal{L}_{nsd}$ is the loss that penalizes the modification of values for pixels outside the shadow mask $M(I)$ for the input image $I$:
$\mathcal{L}_{nsd}(I) = \mymean{i \notin M(I)} \norm{ A(I)_i - I_i }_1$. $\mathcal{L}_{sd}$ is the adversarial loss. It penalizes the correct recognition of D-Net for shadow pixels on the generated image, restricted to the area inside the training shadow mask $M(I)$:  $\mathcal{L}_{sd}(I) = \mymean{i\in {M}(I)}[D(A(I))_i]$. $\mathcal{L}_{ph}$ is a physics-inspired loss to ensure that the shadow area in the generated image is re-illuminated in a physically feasible way. Based on the illumination model described in Section~\ref{sec:illumModel}, we want the ratio $\frac{A(I)_i}{I_i}$ to be similar for all pixels $i$ inside a re-illuminated shadow area. We model this by adding a loss term for the variance of the log ratios 
\begin{align}
\mathcal{L}_{ph}(I) = \sum_{c\in \{R,G,B\}}\myvar{i \in M(I)}\left[\mathrm{log}({A({I})}_i^c) - \mathrm{log}(I_i^c) \right]. \nonumber
\end{align}
where $(\cdot)^c$ denotes the pixel value in the color channel $c$ of the RGB color image.
%
%



\begin{figure}[t] 
\centering
\makebox[0.19\subFigSzab]{}
\makebox[0.19\subFigSzab]{}
\underline{\makebox[0.6\subFigSzab]{Adversarial Examples}}\\
\makebox[0.18\subFigSzab]{(a) Input}
\makebox[0.18\subFigSzab]{(b) GT}
\makebox[0.18\subFigSzab]{(c) Epoch 1}
\makebox[0.18\subFigSzab]{(d) Epoch 5}
\makebox[0.18\subFigSzab]{(e) Epoch 40}
\includegraphics[width=\subFigSzab]{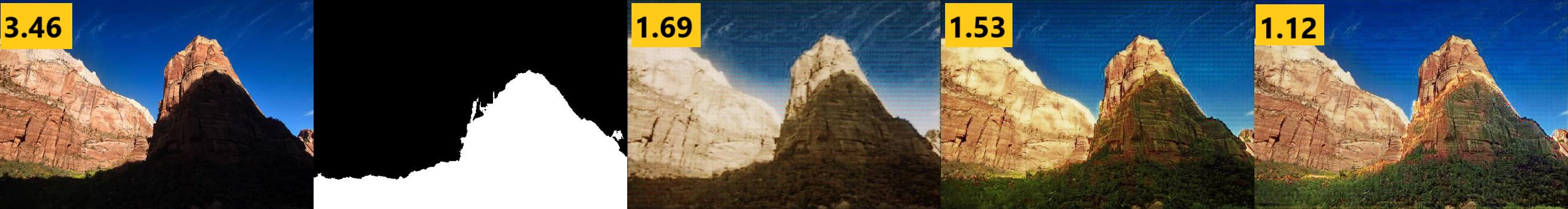}
\includegraphics[width=\subFigSzab]{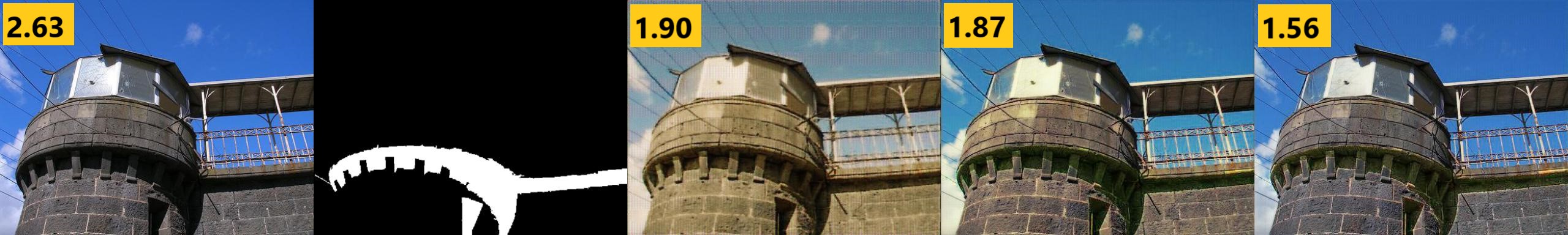}
\vskip -0.1in
\caption{{{\bf Examples of attenuated shadows.}  (a) Input image. (b) Ground truth shadow mask. (c, d, e): adversarial examples with attenuated shadows generated by A-Net from epoch 1, 5, and 40 respectively. The corresponding \textit{shadow strength} are shown as black text on the top-left corner of each image.
}}
\label{fig:attenuatedEx}
\end{figure}

Fig. \ref{fig:attenuatedEx} shows some examples of attenuated shadows that were generated by A-Net during the adversarial training process. The two original input images contain easy to detect shadows with  strengths 3.46 and 2.63. The heuristic to measure these shadow strength values are described in Section \ref{Dnet}. The outputs of A-Net given these input images and shadow masks are shown in columns (c, d, e), obtained at epochs 1, 5, and 40 during  training. The shadows in the generated images become harder to detect as training progresses. Numerically, the shadow strength of the attenuated shadows decreases over time. Moreover, A-Net also learns to not change the non-shadow areas.

\subsection{D-Net: Shadow Detector Network}
\label{Dnet}

The D-Net is central to our framework. It learns to detect shadows from adversarial examples generated by the A-Net as well as original training examples.  On each training iteration, both the original input and the adversarially attenuated image are used to train D-Net. The learning objective for D-Net is to minimize the following loss function:
\begin{align}
\mathcal{L}_D(I) = \  & \lambda_{real}\norm{D(I) - M(I)}_1 + \lambda_{adv}(A(I))\norm{D(A(I)) - M(I)}_1,
\end{align}
where $\lambda_{real}$ and $\lambda_{adv}(A(I))$ control how much D-Net should learn from the real sample $I$ and the adversarial example $A(I)$ respectively. $\lambda_{adv}(A(I))$ depends on how much the shadow in $I$ has been attenuated. If $A(I)$ is the completely shadow-free version of $I$, $\lambda_{adv}(A(I))$ should ideally be zero. Otherwise, this loss function corresponds to having label noise as it requires the output of the shadow detector D-Net for the input $A(I)$ to be the same as the shadow mask $M(I)$, while $A(I)$ is a shadow-free image. 

\begin{figure*}[t]
\centering
\makebox[0.15\subFigSzab]{ Input}
\makebox[0.15\subFigSzab]{ GT Mask}
\makebox[0.15\subFigSzab]{ $\mB_{in}, \mB_{out}$ }
\makebox[0.01\subFigSzab]{   }
\makebox[0.15\subFigSzab]{ Input}
\makebox[0.15\subFigSzab]{ GT Mask}
\makebox[0.15\subFigSzab]{$\mB_{in}, \mB_{out}$}
\includegraphics[width=0.46\subFigSzab,height=0.16\subFigSzab]{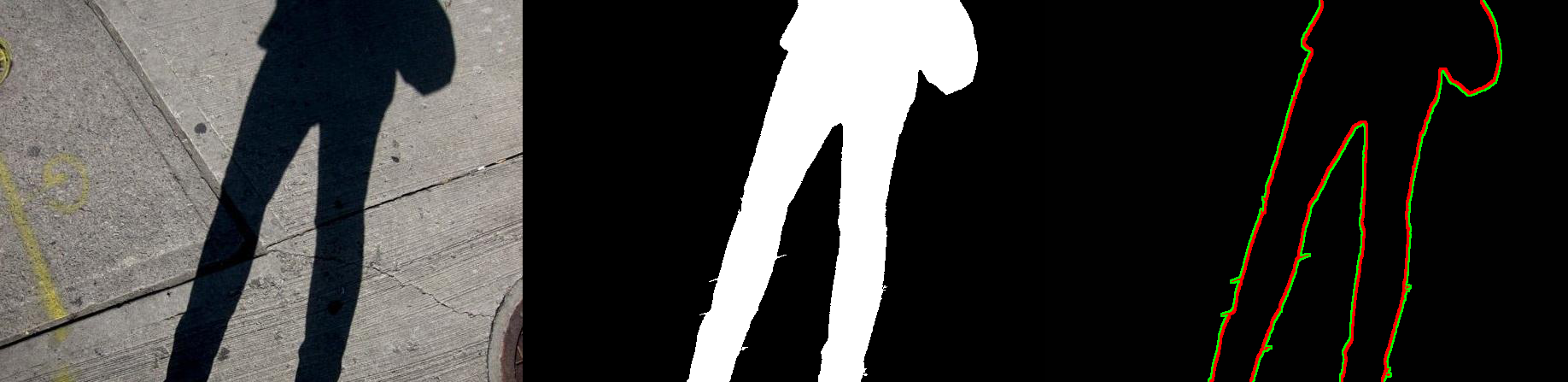} \hspace{0.5cm}
\includegraphics[width=0.46\subFigSzab,height=0.16\subFigSzab]{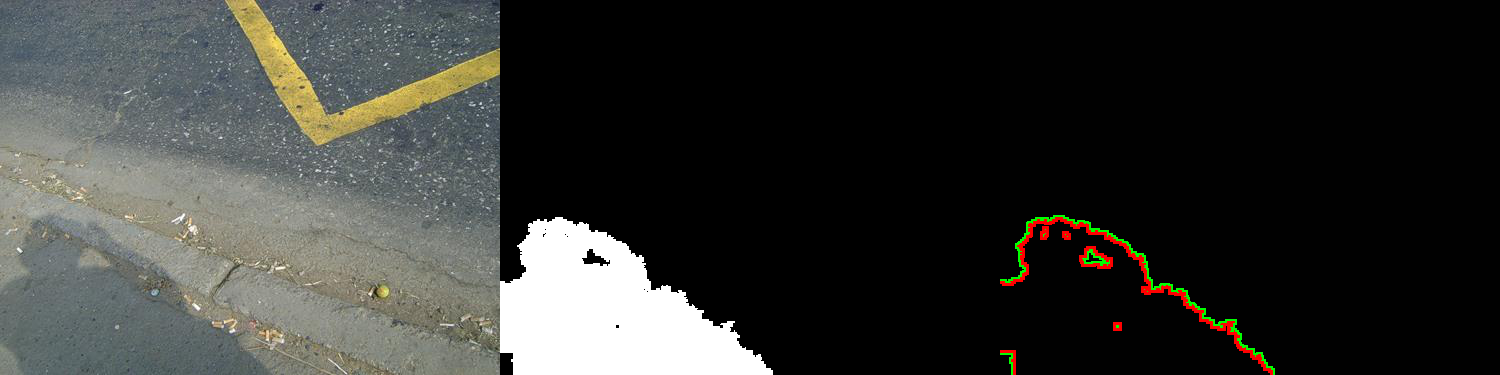} \\
\makebox[0.45\subFigSzab]{(a) Strong shadow ($k_{strength}=4.16$)}
\makebox[0.01\subFigSzab]{   }
\makebox[0.45\subFigSzab]{(b) Weak shadow ($k_{strength}=1.15$)}
\caption{{\bf Estimating the shadow strength.} From the ground-truth shadow mask, we define two area $\mB_{in}$ (red) and $\mB_{out}$ (green) obtained by dilation and erosion of the shadow mask. The shadow strength $k_{strengh}$ is computed as the ratio between the average intensity of pixels in $\mB_{out}$ over the average intensity of pixels in $\mB_{in}$. (a) an image with very a strong dark  shadow, $k_{strength} = 4.16$. (b) light shadow $k_{strength}=1.15$.}
\label{fig:iratio}
\end{figure*}

To determine if $A(I)$ is a shadow-free image, we derive a heuristic based on the illumination model described in Sec.~\ref{sec:illumModel}. We first define two areas alongside the shadow boundary, denoted as $\mB_{in}$ and $\mB_{out}$, illustrated in Fig. \ref{fig:iratio}. $\mB_{out}$ (green) is the area right outside the boundary,  computed by subtracting the shadow mask from its dilated version. The inside area $\mB_{in}$ (red) is computed similarly with the eroded shadow mask. We define the \textsl{shadow strength} $k_{strength}$ as the ratio of average pixel intensities of the two boundary areas: $k_{strength}(A(I))=\frac{\mymean{i \in \mB_{out}}[A(I)_i]}{\mymean{i \in \mB_{in}}[A(I)_i]}$. Fig.  \ref{fig:iratio} shows two examples of images with two different shadow strengths;  an image with a darker shadow (relative to the non-shadow area) has a higher value of $k_{strength}$ and vice versa.

We use the shadow strength of the attenuated image to  decide if D-Net should learn from the attenuated shadow image. Heuristically, the shadow might be completely removed if the shadow strength $k_{strength}$ is too close to 1, i.e., the two areas on the two sides of the shadow boundary have the same average intensities. Based on this heuristic, we set the weight for the adversarial example $A(I)$ as follows: 
\begin{align}
	\lambda_{adv}(A(I)) = \left\{ 
	\begin{array}{ll}
		\lambda_{adv}^{0} & \hspace{3ex}\textrm{if } {k_{strength}(A(I))} > 1 + \epsilon \\
		0 & \hspace{3ex}\textrm{otherwise},
	\end{array}
	\right. 
\end{align}
where $\lambda_{adv}^{0}$ is a tunable baseline factor for adversarial examples and $\epsilon$ is a small threshold which we empirically set to 0.05.

\subsection{Network Architectures}

Both A-Net and D-Net were developed based on the U-Net architecture~\cite{UNet}. Following~\cite{isola2017image}, we created networks with seven skip-connection modules, each of which contains a sequence of Convolutional, BatchNorm, and Leaky-ReLu \cite{LReLu} layers. The A-Net input is a four channel image, which is the concatenation of the RGB image and the corresponding shadow mask. The A-Net output is a three channel RGB image. The input to D-Net is an RGB image, and the output is a single channel shadow mask. 

\section{Experiments and Results \label{sec:exp}}

We experiment on several public shadow datasets. One of them is the SBU Shadow dataset~\cite{Vicente-etal-ECCV16}. 
 This dataset consists of pairs of RGB images and corresponding annotated shadow binary masks. The SBU dataset contains 4089 training images, and 638 testing images, and is currently the largest and most challenging shadow benchmark. We also perform cross-dataset experiments on the UCF testing set~\cite{Zhu10}, which contains 110 images with corresponding shadow masks.
We quantitatively evaluate shadow detection performance by comparing the testing ground-truth shadow masks with the prediction masks produced by D-Net. As is common practice in the shadow detection literature, we will use the Balanced Error Rate (BER) as the  principal evaluation metric. The BER is defined as:
$\mbox{BER} = 1 - \frac{1}{2} \left( \frac{TP}{TP+FN} + \frac{TN}{TN+FP} \right)
$, where $TP, TN, FP, FN$ are the total numbers of true positive, true negative, false positive, and false negative pixels respectively. Since natural images tend to  overwhelmingly more  non-shadow pixels, the BER is less biased than mean pixel accuracy. We also provide separate mean pixel error rates for the shadow and non-shadow classes.

\myheading{Training and implementation details.} 
We use stochastic gradient descent with the Adam solver~\cite{Adam} to train our model. We use mini batch SGD with batch size of 64. On each training iteration, we perform three forward passes consecutively: forward the input shadow image $I$ to A-Net to get the adversarial example $A(I)$, then separately forward the adversarial image and shadow input image to D-Net. We alternate one parameter update step on D-Net with one update step on A-Net, as suggested by~\cite{GoodfellowPMXWOCB14}. Before training and testing, we transform the images into log-space. We experimentally set our training parameters as:
$(\lambda_{nsd},\lambda_{sd},\lambda_{ph},\lambda_{real},\lambda_{adv}^{0}):=(30,1,100,0.8,0.2)$. We implemented our framework on PyTorch. More details can be found at: \url{www3.cs.stonybrook.edu/~cvl/projects/adnet/index.html}

\subsection{Shadow Detection Evaluation }
We evaluate the shadow detection performance of the proposed D-Net on the SBU and UCF datasets. To detect shadows in an image, we first resize the image to $256\times256$. We input this image to D-net to produce a shadow mask of size $256\times 256$, which will be   compared with the ground-truth shadow mask for evaluation (in the original size).

\setlength{\tabcolsep}{5pt}

In Table \ref{table:quant}, we compare the performance of our method with the state-of-the-art methods Stacked-CNN~\cite{Vicente-etal-ECCV16}, scGAN~\cite{VuICCV2017}, ST-CGAN~\cite{Wang_2018_CVPR}, and DSC~\cite{Hu_2018_CVPR}. We also consider a variant of D-Net, trained without the attenuated shadow images from A-Net. All methods are trained on the SBU training set. Performance is reported in terms of BER, as well as shadow and non-shadow error rates. Note that DSC~\cite{Hu_2018_CVPR} only reported BER numbers on the SBU dataset and its cross-domain results were obtained on testing data that is different from the commonly used UCF test dataset (as proposed by~\cite{Zhu10}).

On the SBU test set, our detector (D-Net) outperforms the previous state-of-the-art methods. Compared to the Stacked-CNN we obtain a 51\% error reduction. 
Compared to scGAN and ST-CGAN, D-Net brings a 41\% error reduction and a 33\% error reduction respectively. D-Net outperforms DSC by 0.2\% BER, aeven though it is significantly simpler. D-Net is fully convolutional, without the need of for running recurrent neural networks and CRF post processing. 


For the cross-dataset experiments, the detectors are trained on the SBU training set, but they are evaluated on the test set of the UCF dataset~\cite{Zhu10}. These datasets are disjoint; while SBU covers a wide range of scenes, UCF focuses on images where dark shadows as well dark albedo objects are present. Again, we compare our method with the previous state-of-the-art methods: Stacked-CNN~\cite{Vicente-etal-ECCV16}, scGAN~\cite{VuICCV2017}, and ST-CGAN~\cite{Wang_2018_CVPR}. In terms of BER, our proposed D-Net yields significant error reductions of 18\% and 16\% with respect to scGAN and ST-CGAN, respectively. 
The performance gap between D-Net trained with and without attenuated shadow images is very significant, highlighting the benefits of having attenuated shadow examples for training.
\setcounter{footnote}{0}
\begin{table}[th]
\centering
\caption{{\bf Evaluation of shadow detection methods on the SBU Shadow dataset~\cite{Vicente-etal-ECCV16} and for  cross-dataset detection on UCF~\cite{Zhu10}}. All methods are trained on the SBU training data. Both Balanced Error Rate (BER) and per class error rates are shown. DSC~\cite{Hu_2018_CVPR} only reported BER numbers, and used a different UCF test dataset, so  cross-domain performance cannot be compared.  
Best performances is printed in bold.}
\vskip 0.1in
\label{table:quant}
\begin{tabular}{lcccccc}
\toprule
& \multicolumn{3}{c}{Evaluated on SBU Testset~\cite{Vicente-etal-ECCV16}}                   & \multicolumn{3}{c}{Evaluated on UCF Testset~\cite{Zhu10}}                    \\ 
\cmidrule(lr){2-4}  
\cmidrule(lr){5-7} 
Method & BER          & Shadow       & Non Shad.    & BER          & Shadow       & Non Shad.     \\ 
\midrule
stacked-CNN~\cite{Vicente-etal-ECCV16} & 11.0         & 9.6          & 12.5         & 13.0         & 9.0          & 17.1          \\
scGAN~\cite{VuICCV2017}                & 9.1          & 7.8          & 10.4         & 11.5         & 7.7          & 15.3          \\ 
ST-CGAN~\cite{Wang_2018_CVPR} 	  & 8.1          & \textbf{3.7}          & 12.5         & 11.2         & \textbf{5.0}          & 17.5          \\
DSC~\cite{Hu_2018_CVPR} 	  & 5.6          &-          & -         &-        & -          & -          \\
\hline
D-Net (w/o A-Net)                       & 8.8          & 8.1          & 9.3          & 11.8         & 8.9          & 14.7          \\
D-Net (with A-Net)                             & \textbf{5.4} & 5.3 & \textbf{5.5} & \textbf{9.4} & 7.0 & \textbf{11.8} \\ \hline
\end{tabular}
\end{table}

\FloatBarrier
\subsection{Qualitative Results}
\FloatBarrier


\begin{figure*}[ht]
\centering

\makebox[0.15\subFigSzab]{(a) Input}
\makebox[0.15\subFigSzab]{(b) GT}
\makebox[0.15\subFigSzab]{(c) Ours} 
\makebox[0.15\subFigSzab]{(a) Input}
\makebox[0.15\subFigSzab]{(b) GT}
\makebox[0.15\subFigSzab]{(c) Ours} \\
\includegraphics[width=0.45\subFigSzab,height=0.16\subFigSzab]{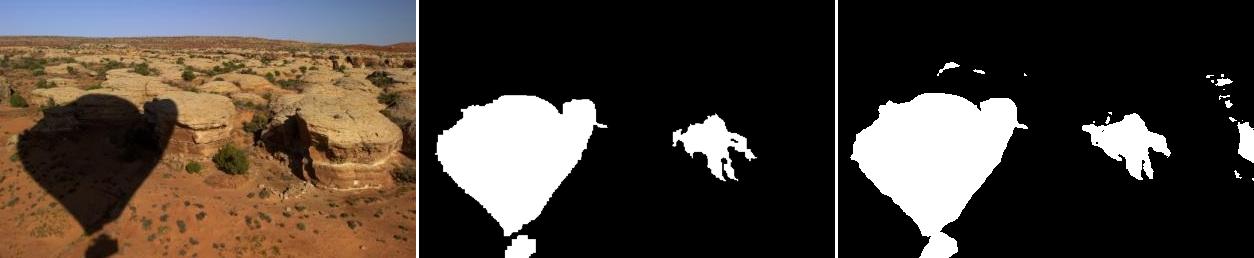} 
\includegraphics[width=0.45\subFigSzab]{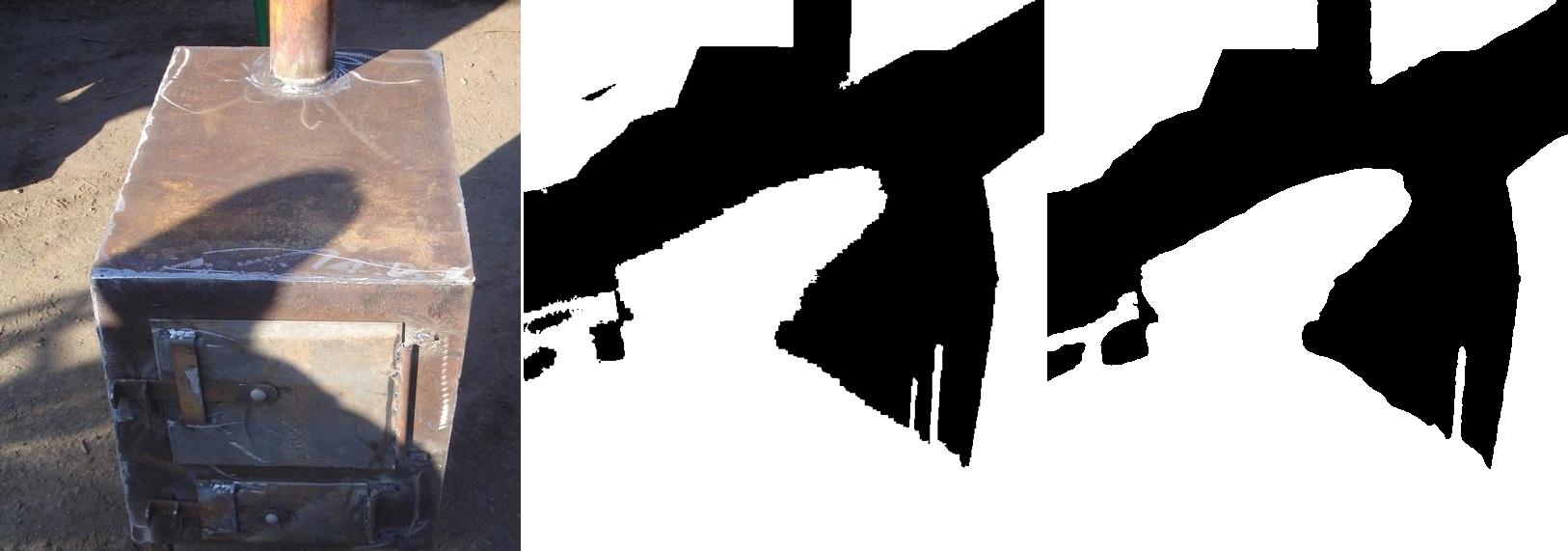} \\
\includegraphics[width=0.45\subFigSzab]{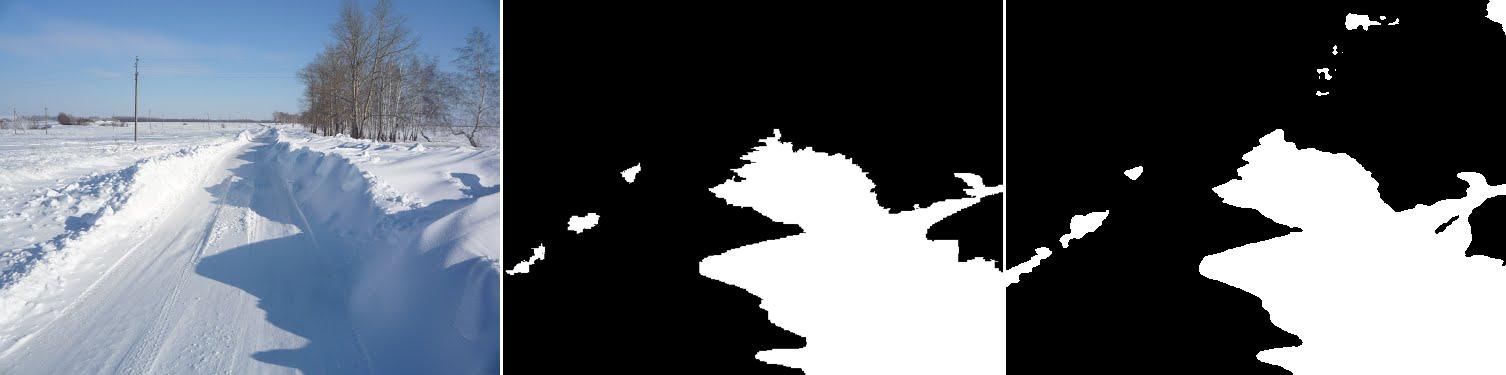} 
\includegraphics[width=0.45\subFigSzab]{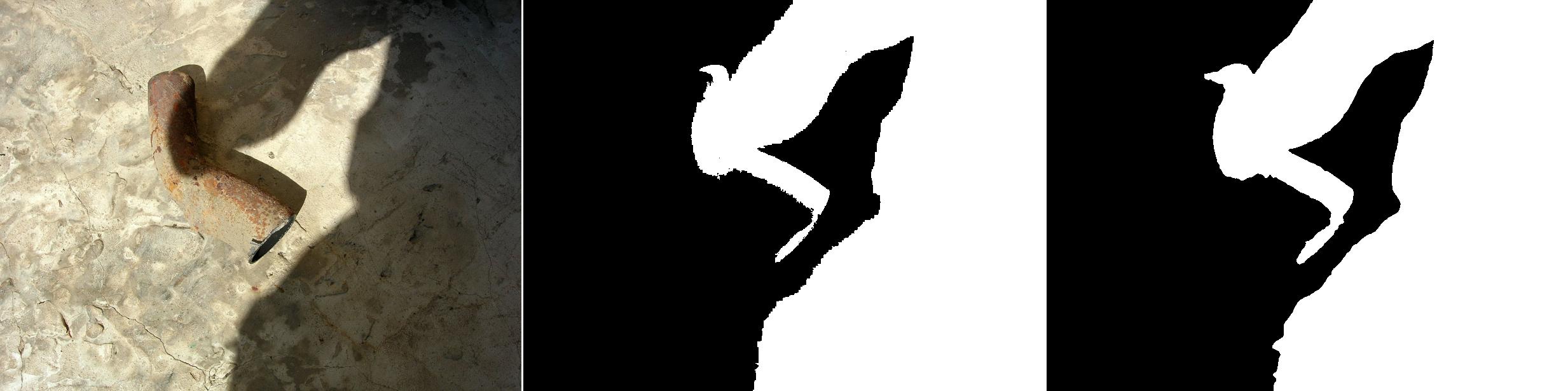} \\
\includegraphics[width=0.45\subFigSzab,height=0.14\subFigSzab]{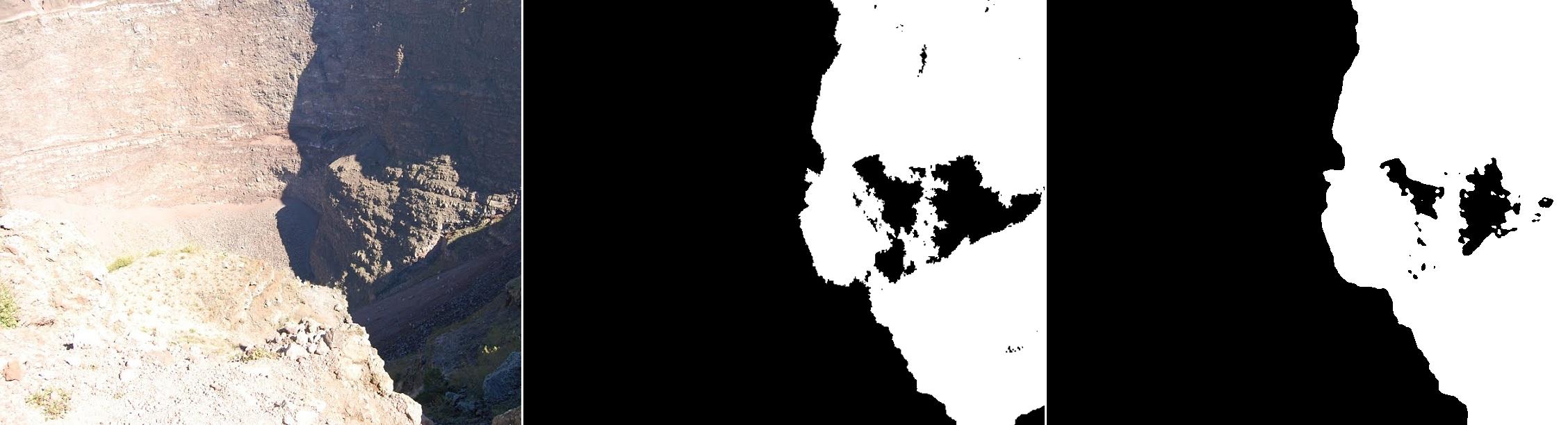} 
\includegraphics[width=0.45\subFigSzab]{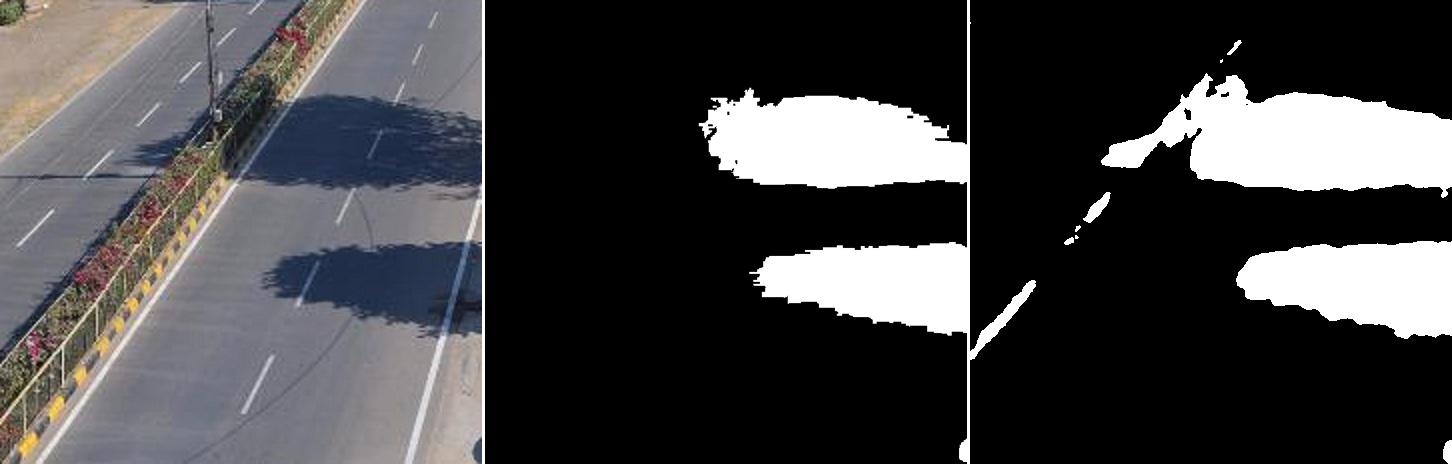} \\
\includegraphics[width=0.45\subFigSzab,height=0.11\subFigSzab]{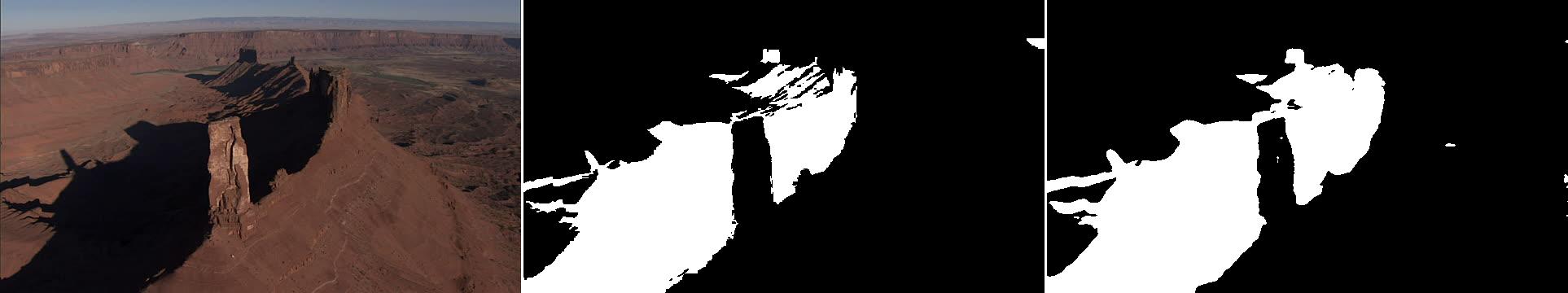} 
\includegraphics[width=0.45\subFigSzab]{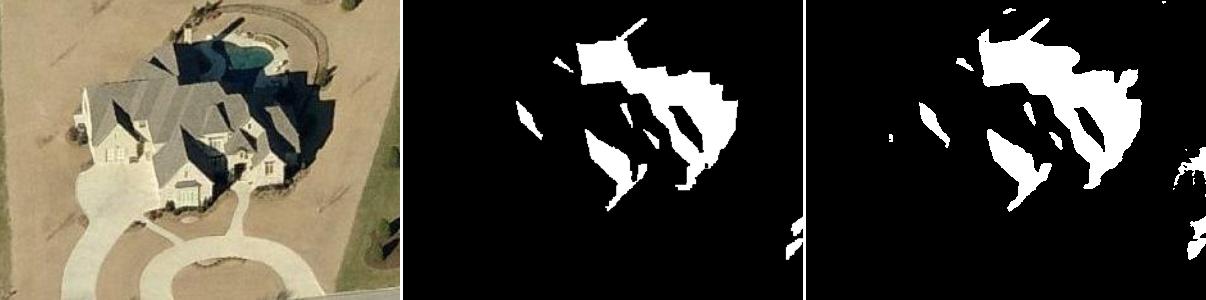}  \\
\makebox[0.45\subFigSzab]{(i)}
\makebox[0.45\subFigSzab]{(ii)}
\vskip -0.1in 
\caption{{{\bf Shadow detection results.} Our proposed method accurately detects shadows on: (i) different scenes, and illumination conditions; (ii) close-ups  and long-range shots, as well as aerial images.}}
\label{fig:impress_1and2}
\end{figure*}

In Fig. \ref{fig:impress_1and2} (i) and (ii), we show shadow detection results on the SBU dataset. The columns show input images,  ground truth shadow masks, and  D-Net outputs, respectively. In Fig. \ref{fig:impress_1and2}.(i), we see how  the D-Net correctly predicts shadows on different types of scenes such as desert, mountain, snow, and under different weather conditions from sunny to cloudy and overcast. In Fig. \ref{fig:impress_1and2}.(ii), notice how the D-Net accurately predicts shadows in close-ups as well as long-range shots, and in aerial images. Fig. \ref{fig:compare} shows qualitative comparisons with the shadow detection results of scGAN~\cite{VuICCV2017}. In general, D-Net produces more accurate shadows with sharper boundaries.  

\def\subFigSzab{\linewidth}
\def\highsub{1.8cm}

\begin{figure}
\centering
\makebox[0.23\subFigSzab]{(a) Input}
\makebox[0.23\subFigSzab]{(b) GT}
\makebox[0.23\subFigSzab]{(c) scGAN}
\makebox[0.23\subFigSzab]{(d) Ours}
\includegraphics[width=0.9\subFigSzab,height=\highsub]{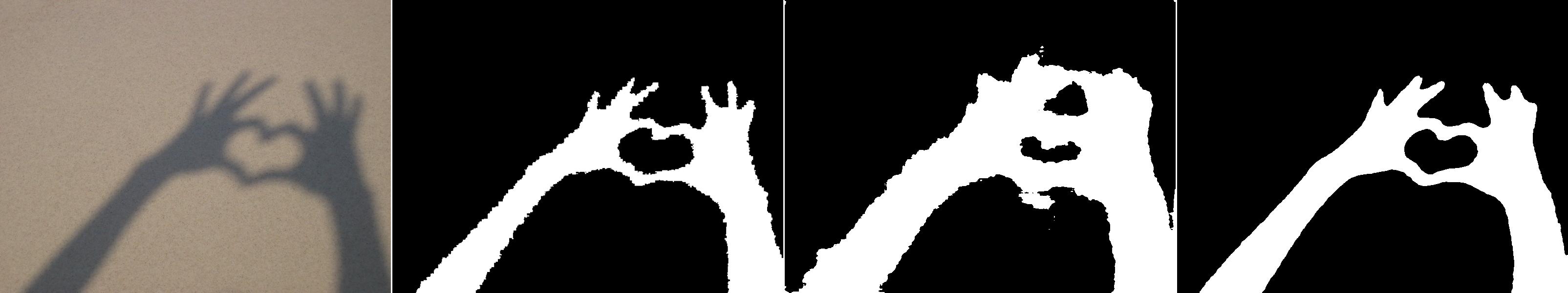}
\includegraphics[width=0.9\subFigSzab,height=\highsub]{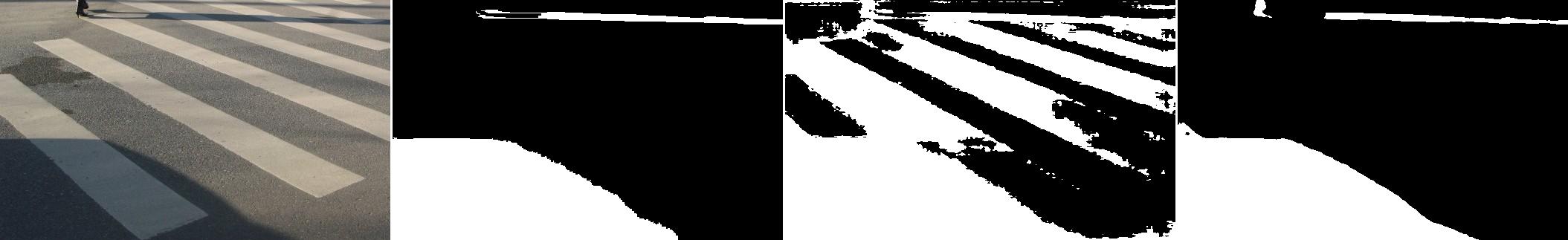}
\includegraphics[width=0.9\subFigSzab,height=\highsub]{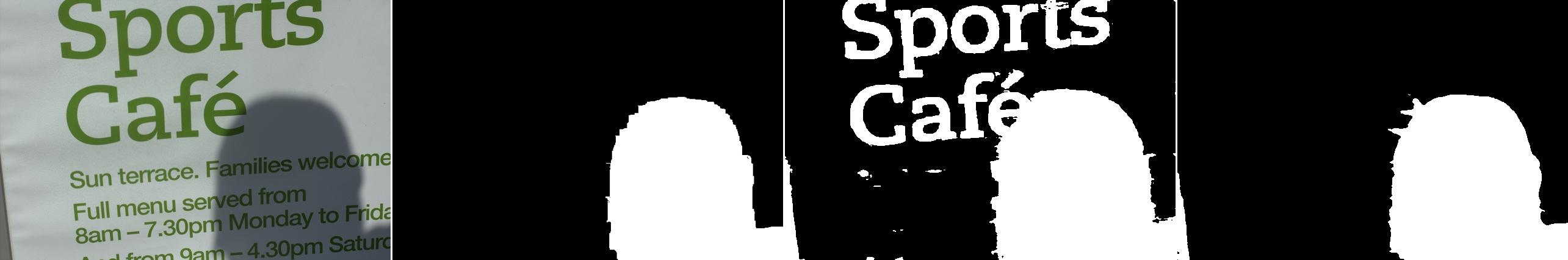}

\caption{{{\bf Comparison of shadow detection on SBU dataset.} Qualitative comparison between our method and the state-of-the-art method scGAN~\cite{VuICCV2017}. (a) Input image. (b) Ground-truth shadow mask. (c) Predicted shadow mask by scGAN~\cite{VuICCV2017}. (d) Predicted shadow mask by our method.}}
\label{fig:compare}
\end{figure}

\def\subFigSzab{\linewidth}
\begin{figure}[ht!] 
\centering
\begin{minipage}{.36\textwidth}
\centering
\makebox[0.3\subFigSzab]{(a) Input}
\makebox[0.3\subFigSzab]{(b) GT}
\makebox[0.3\subFigSzab]{(c) Ours}
\includegraphics[width=\subFigSzab]{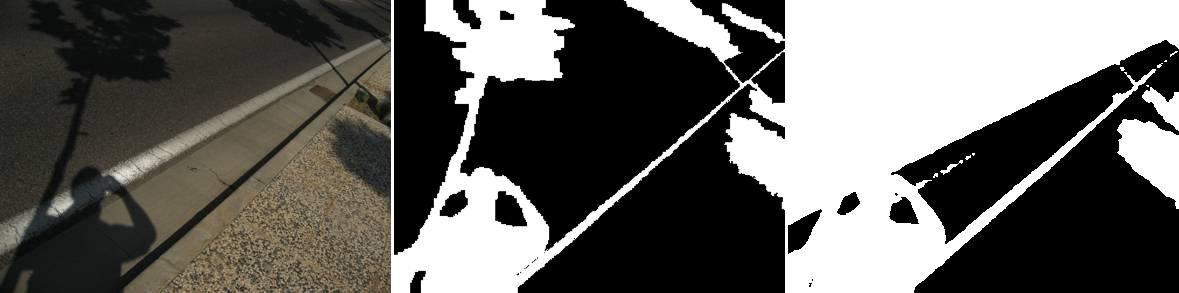}
\includegraphics[width=\subFigSzab]{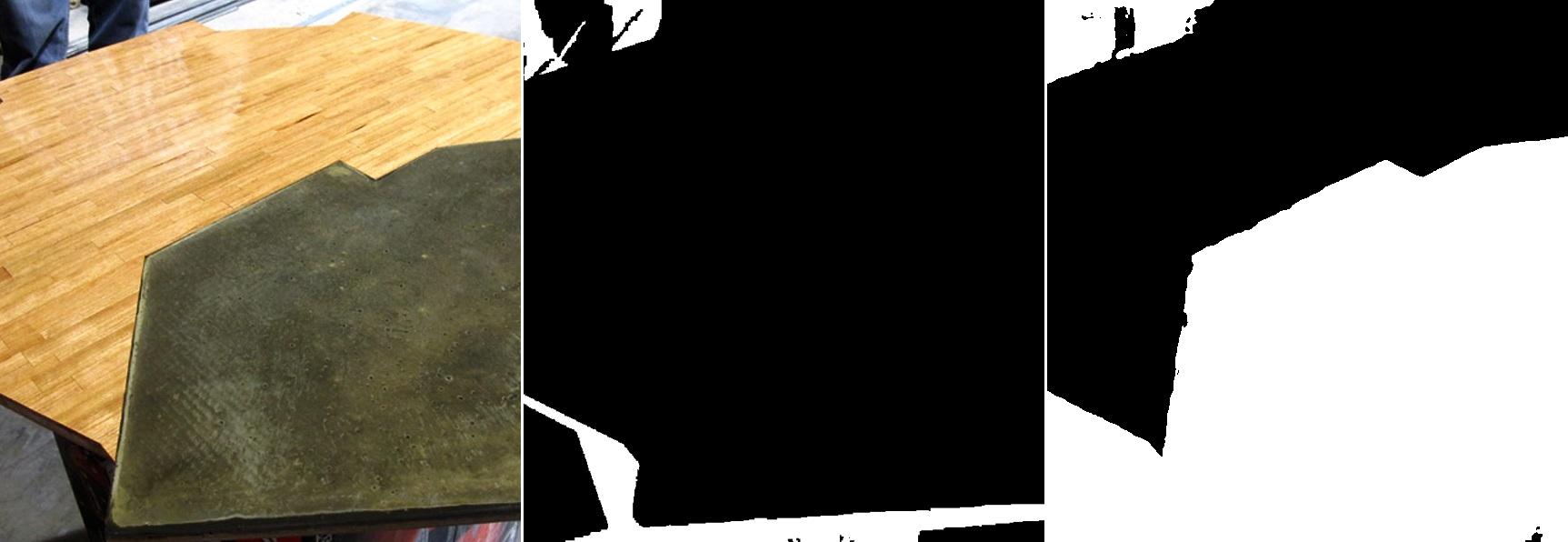}
\def\subFigSzab{\linewidth}
\caption{{{\bf Failed shadow detection examples.} Failure cases of our method due to non-shadow dark albedo regions. (a) Input image. (b) Ground-truth mask. (c) Predicted shadow mask by our method.}}
\label{fig:fail}
\end{minipage}%
\hspace{0.5cm}
\begin{minipage}{.5\textwidth}
\centering
\includegraphics[width=\subFigSzab]{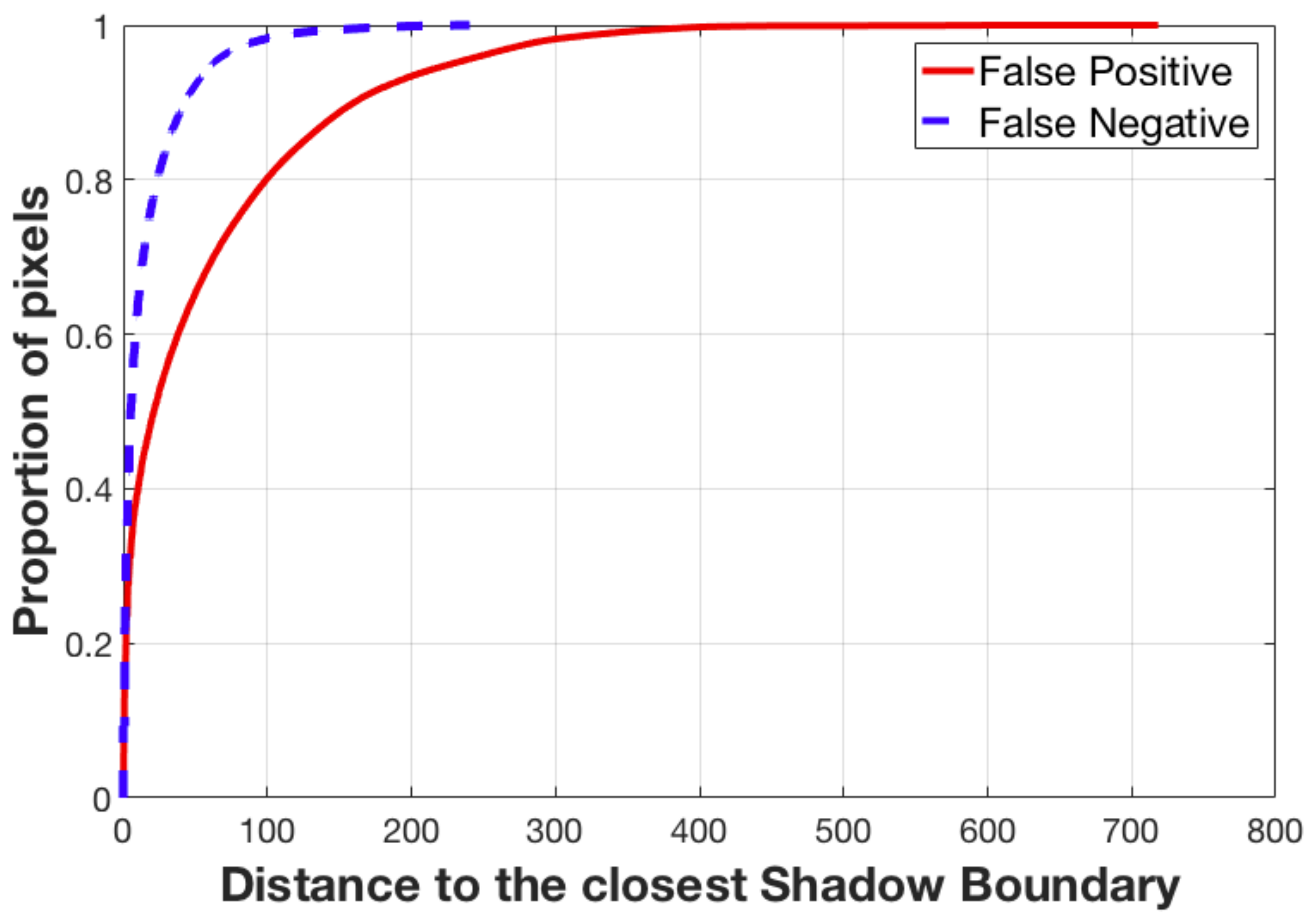}
\caption{{{\bf Cumulative curve of the distance of wrongly predicted pixels to the closest shadow boundary on the SBU testing set.}}}
\label{fig:cummulative_fail}
\end{minipage}%
\end{figure}


\subsection{Failure Cases}

Some failure cases of our method are shown in Fig. \ref{fig:fail}. Many are due to dark albedo material regions being incorrectly classified as shadows. We also investigate the locations of wrongly classified pixels to understand the causes of failure. Fig. \ref{fig:cummulative_fail} shows the proportion of wrongly predicted pixels with respect to their distances to the closest ground-truth shadow boundary on the SBU testing set. A large portion of missed shadow pixels is within a small distance to a boundary. Specifically, 65\% of false negative cases are within 10 pixels of a shadow boundary. This means the shadow pixels missed by our method are probably either around the shadow boundaries or inside very small shadow regions. Meanwhile a large portion of false positive prediction is far away from a shadow boundary. This is perhaps due to the misclassifications of dark objects as shadows.


\subsection{Ablation Study and Parameter Analysis }
\setlength{\tabcolsep}{3pt}
\begin{table}[h]
\centering
\caption{\textbf{Ablation study.} Comparison of shadow detection results of our framework  with and without inclusion of the physics based loss $L_{ph}$. Detection performance significantly profits from incorporating the physics based loss $L_{ph}$ into the training process: 20\% reduction of BER in SBU\cite{Vicente-etal-ECCV16} testing set, and 27\% error reduction in UCF~\cite{Zhu10}~(cross-dataset task)}
\label{table:ablation}
\vskip 0.1in
\begin{tabular}{lcccccc}
\toprule
& \multicolumn{3}{c}{Evaluated on SBU Testset} & \multicolumn{3}{c}{Evaluated on UCF Testset}                    \\ 
\cmidrule(lr){2-4}  
\cmidrule(lr){5-7} 
Method                        & BER & Shadow  & Non Shad.    & BER          & Shadow       & Non Shad.     \\ 
\midrule
D-Net ($+L_{ph}$, $+\lambda_{adv}$)   & \textbf{5.4} & \textbf{5.3} & 5.5          & \textbf{9.4} & \textbf{7.0} & \textbf{11.8} \\ 

D-Net ($+L_{ph}$, $-\lambda_{adv}$)  & 5.7          & 6.2          & \textbf{5.2} & 9.9          & 7.3          & 12.5          \\ 
D-Net ($-L_{ph}$,  $-\lambda_{adv}$) & 7.1          & 7.6          & 6.7          & 13.6         & 15.9         & 11.3          \\
\bottomrule
\end{tabular}
\end{table}

\def\subFigSzab{\linewidth}
\begin{figure} 
\centering
\makebox[0.3\subFigSzab]{(a) Input}
\makebox[0.3\subFigSzab]{(b) Result w/o $L_{ph}$}
\makebox[0.3\subFigSzab]{(c) Result w/ $L_{ph}$}
\includegraphics[width=0.8\subFigSzab]{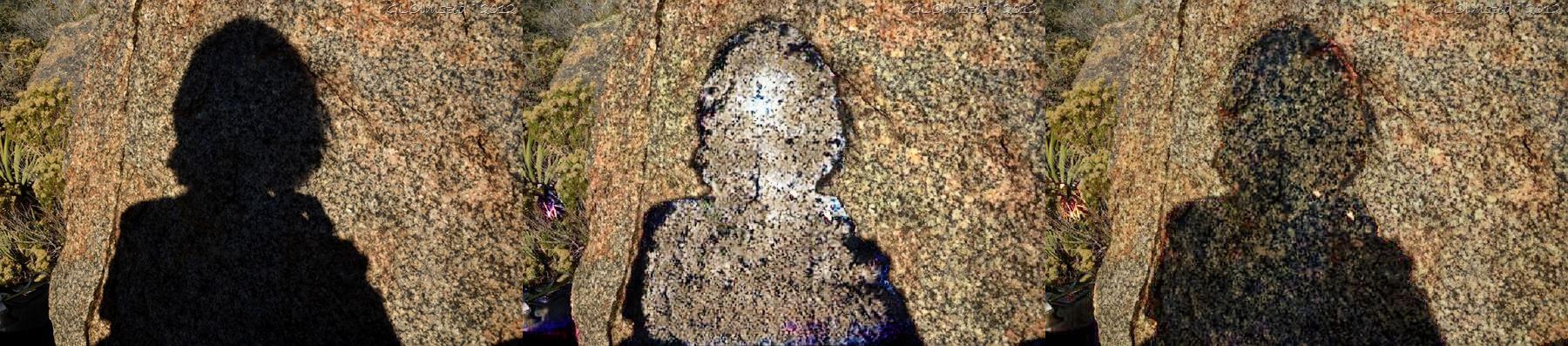}
\includegraphics[width=0.8\subFigSzab]{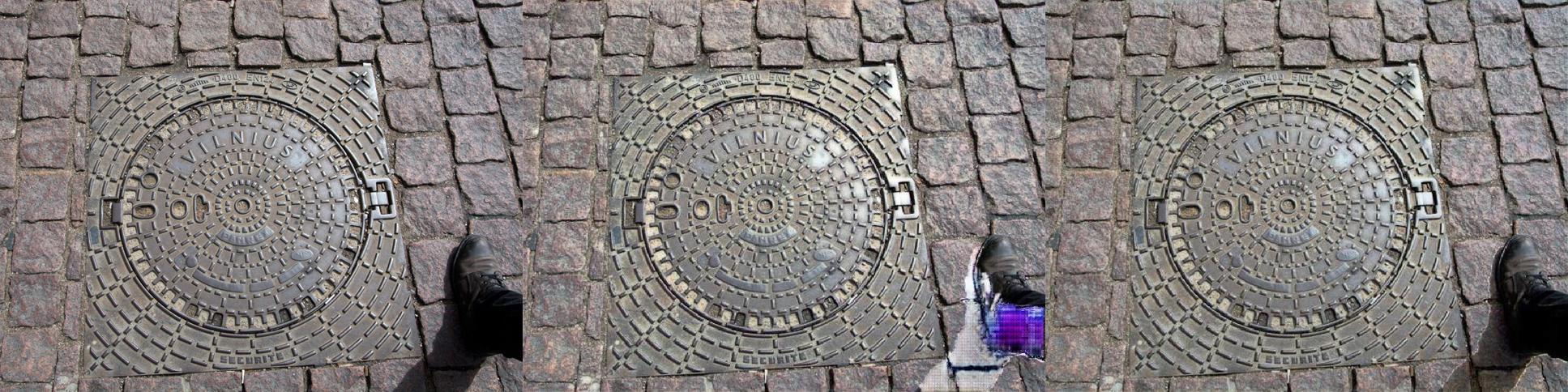}
\caption{{{\bf Examples of adversarial examples generated with and without physics.} (a) Input image \textit{I}. (b) Adversarial example generated by A-net trained without physics based loss. (c) Adversarial example generated by A-net trained with physics based loss.}}
\label{fig:phy}
\end{figure}

We conducted experiments to analyze the impact of the physics-based loss ($L_{ph}$) and the weight function $\lambda_{adv}$ in our framework. We trained our model with two additional scenarios for comparison: 1) without the physics-based loss and without the weight function $\lambda_{adv}$, and 2) with the physics-based loss but without the weight function $\lambda_{adv}$. We denote these two configurations as ($-L_{ph}$,  $-\lambda_{adv}$) and $(+L_{ph}$,  $-\lambda_{adv}$) respectively.
Table \ref{table:ablation} shows the shadow detection results of the models trained with these modified conditions. We tested the models, trained on SBU, on both the UCF and SBU testing sets. As can be seen from Table \ref{table:ablation}, dropping the weight function $\lambda_{adv}$ increased error rates slightly, while dropping the physics-based loss drastically increased error rates. In Fig. \ref{fig:phy}, we compare adversarial examples generated by the model trained with and without the physics-based loss. Incorporating this loss produces images with more realistic attenuated shadows. Thus, the produced examples aid the  training of the shadow detector D-Net. In our experiments, at the $50^{th}$ training epoch,  approximately 6\%  of all  images  generated by A-Net, were not used based on $\lambda_{adv}$.

We conducted experiments to study the effect of the parameters of our framework. We started from the parameter settings reported in Section \ref{sec:exp}. When we chose $\lambda_{sd} = 10$, D-Net achieved 6.5\% BER. As $\lambda_{sd}$ increases, A-Net attenuates the shadow more dramatically but also tends to change the non-shadow part, generating lower quality images in general. In the second experiment, we rescaled the ratio between the real and adversarial images being input to D-Net. When we chose $\lambda_{adv}^{0}  = 0.5$ and $\lambda_{real} = 0.5$, D-Net achieved 7.0\% BER. \\

\section{Summary}
In this paper, we have presented a novel framework for adversarial training of a shadow detector using shadow attenuation. We have shown experimentally how our model is able to effectively learn from both real shadow training examples as well as adversarial examples. Our trained model outperforms the previous state-of-art  shadow detectors in two benchmark datasets, demonstrating the effectiveness and generalization ability of our model. Furthermore, to the best of our knowledge, this is the first shadow detector that can detect shadows accurately at real-time speed, 45 fps. 

\myheading{Acknowledgements.} This work was supported by the Vietnam Education Foundation, a gift from Adobe, NSF grant CNS-1718014, the Partner University Fund, and the SUNY2020 Infrastructure Transportation Security Center. The authors would also like to thank NVIDIA for GPU donation. 
\FloatBarrier
{\small
\bibliographystyle{splncs04}
\bibliography{longstrings,egbib}
}

\end{document}